\documentclass{article}
\usepackage[preprint]{main}
\usepackage[utf8]{inputenc} 
\usepackage[T1]{fontenc}    
\usepackage{hyperref} 
\hypersetup{hidelinks}
\usepackage{url}      
\usepackage{booktabs}       
\usepackage{amsfonts}       
\usepackage{nicefrac}       
\usepackage{microtype}     
\usepackage{xcolor}   
\usepackage{enumitem}
\usepackage{multirow}
\usepackage{graphicx}
\usepackage{caption}
\usepackage{floatrow}
\newfloatcommand{capbtabbox}{table}[][\FBwidth]
\usepackage{graphicx}
\usepackage{marvosym}
\usepackage{natbib}

\title{TextSquare: Scaling up Text-Centric Visual Instruction Tuning}

\author{
 \textnormal{
 \normalsize{Jingqun Tang\textsuperscript{\rm 1}$^{*}$, Chunhui Lin\textsuperscript{\rm 1}$^{*}$, Zhen Zhao\textsuperscript{\rm 2}$^{*}$, Shu Wei\textsuperscript{\rm 1}, Binghong Wu\textsuperscript{\rm 1}, Qi Liu\textsuperscript{\rm 1}, Yangfan He\textsuperscript{\rm 4},}
} 
\\
\normalsize{Kuan Lu\textsuperscript{\rm 5}, Hao Feng\textsuperscript{\rm 1},
Yang Li\textsuperscript{\rm 1}, Siqi Wang\textsuperscript{\rm 1}, Lei Liao\textsuperscript{\rm 1}, Wei Shi\textsuperscript{\rm 1}, Yuliang Liu\textsuperscript{\rm 3}, Hao Liu\textsuperscript{\rm 1}, }
\\
\normalsize{
Yuan Xie\textsuperscript{\rm 2},Xiang Bai\textsuperscript{\rm 3}, Can Huang\textsuperscript{\rm 1 \Letter}}
\\
\textnormal{\small{\textsuperscript{\rm 1}ByteDance Inc. \textsuperscript{\rm 2} East China Normal University \textsuperscript{\rm 3}Huazhong University of Science and Technology}}
\\
\textnormal{\small{\textsuperscript{\rm 4}University of Minnesota \textsuperscript{\rm 5} Cornell University }}
\\
\small{tangjingqun@bytedance.com, can.huang@bytedance.com}
}

\begin{document}

\maketitle

\begin{abstract}

Text-centric visual question answering (VQA) has made great strides with the development of Multimodal Large Language Models (MLLMs), yet open-source models still fall short of leading models like GPT4V and Gemini, partly due to a lack of extensive, high-quality instruction tuning data.  
To this end, we introduce a new approach for creating a massive, high-quality instruction-tuning dataset, Square-10M, which is generated using closed-source MLLMs.
The data construction process, termed Square, consists of four steps: \textbf{S}elf-\textbf{Qu}estioning, \textbf{A}nswering, \textbf{R}easoning, and \textbf{E}valuation. 
Our experiments with Square-10M led to three key findings: 
1) Our model, TextSquare, considerably surpasses open-source previous state-of-the-art Text-centric MLLMs and sets a new standard on OCRBench (62.2\%). It even outperforms top-tier models like GPT4V and Gemini in 6 of 10 text-centric benchmarks. 2) Additionally, we demonstrate the critical role of VQA reasoning data in offering comprehensive contextual insights for specific questions. This not only improves accuracy but also significantly mitigates hallucinations.  Specifically, TextSquare scores an average of 75.1\% across four general VQA and hallucination evaluation datasets, outperforming previous state-of-the-art models. 3) Notably, the phenomenon observed in scaling text-centric VQA datasets reveals a vivid pattern:  the exponential increase of instruction tuning data volume is directly proportional to the improvement in model performance, thereby validating the necessity of the dataset scale and the high quality of Square-10M. 

\end{abstract}

\renewcommand{\thefootnote}{}
\footnotetext{$^{*}$ Equal contribution. \Letter Corresponding author.}

\section{Introduction}

\begin{figure}[htbp]
\centering
\includegraphics[width=\textwidth]{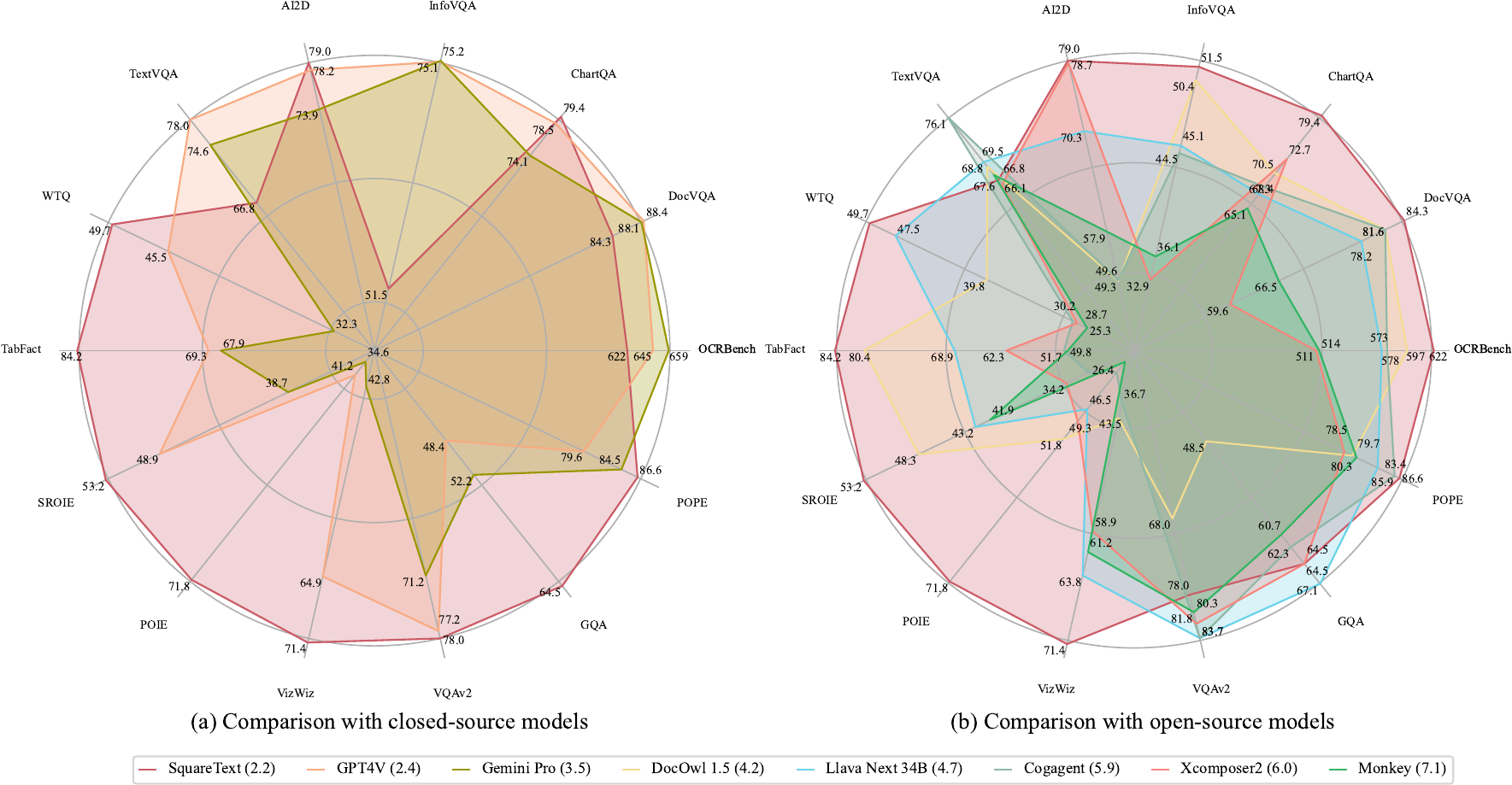}
\caption{ The performance of TextSquare in various VQA tasks compared to existing models. (a) shows the comparison with state-of-the-art closed-source models (Gemini \cite{gemini-pro} and GPT4V \cite{gpt4v}), and (b) shows the comparison with the leading open-source models. The numbers in parentheses after the model names in the legend indicate the average performance ranking across 10 text-centric multimodal benchmarks. TextSquare is marginally superior to GPT4V. Best viewed on screen.}
\label{img-performance}
\end{figure}

Recent research on multimodal large language models (MLLMs) has achieved significant advancements in the text-centric visual question-answering(VQA) domain \cite{Text-MLLM-1,Text-MLLM-2,Text-MLLM-3,docpedia,zhao2025tabpedia, lu2024bounding,zhao2024harmonizing, zhao2024multi}, with several closed-source state-of-the-art (SOTA) models leading the way. Two representative examples are GPT4V \cite{gpt4v} and Gemini \cite{gemini-pro}, which have demonstrated remarkable performance and have even surpassed human-level capabilities in certain aspects. Nevertheless, as illustrated in Figure \ref{img-performance}, the performance of open-source models still lags significantly behind that of pioneering closed-source models. This phenomenon can be attributed to various factors, including model architecture, the scale of model parameters, image resolution, the volume of pretraining and instruction tuning data, and training strategies, among others.

Many pioneering studies \cite{allava, bonito,sharegpt4v,llavar,wang2025pargo,wang2025mari,li2024falcon,sun2025attentive} have recently conducted data-centric research into the challenges of insufficient instruction tuning data. For instance, Monkey \cite{monkey} initially employed expert models to generate descriptions of different aspects of images, which were then summarized by GPT-4 to produce high-quality and detailed image caption data. For better text-based knowledge injection,  For better text-based knowledge injection, LLaVAR \cite{llavar} and TG-Doc \cite{tg-doc} used GPT-4 to generate conversations for text-rich images by integrating OCR \cite{liu2023spts,tang2022optimal,tang2022few,zhao2024multi,tang2023character, tang2024mtvqa,feng2023unidoc,feng2024docpedia,fu2024ocrbench} results into the instructions. In order to improve the image caption ability for MLLMs, ShareGPT4V \cite{sharegpt4v} constructs a high-quality image caption dataset through GPT4V. While these efforts have achieved remarkable success, they have also left some challenges unresolved. Image caption data and VQA data belong to different domains, with inconsistencies in the granularity and scope of image content presentation. Furthermore, the scale of synthetic data remains relatively small, preventing MLLMs from fully realizing their potential. The exploration of methods that leverage large-scale text-centric VQA data for instruction tuning of existing open-source models remains limited.

To bridge the gap, this paper proposes a strategy termed Square for obtaining massive, high-quality text-centric VQA data from sophisticated and versatile closed-source MLLMs, resulting in the construction of a dataset (Square-10M) comprising tens of millions of instances for instruction tuning. Specifically, the method consists of four steps: Self-Questioning, Answering, Reasoning, and Evaluation. The self-questioning step involves utilizing the MLLM's capabilities in text-image analysis and understanding to generate questions related to the textual content of images. The answering step involves answering these generated questions, leveraging various prompting techniques such as Chain-of-Thought and few-shot prompting. The reasoning step entails probing the model for the reasoning behind its answers, leveraging the powerful reasoning abilities of MLLMs. The evaluation step involves evaluating the question-answer pairs, assessing the validity of the questions and their relevance to the textual content of the images, as well as the correctness of the answers, thereby improving data quality and mitigating hallucinations. Overall, Square comprehensively leverages the capabilities of MLLMs in various aspects, significantly enhancing the data quality.

Besides, enriching the diversity of images is also crucial. We collect a diverse set of text-rich images from various public sources, including natural scenes, charts, tables, receipts, books, slides, PDFs, documents, products, and web images. Subsequently, deduplication is performed on this collection. 
By applying the Square method to these images, Square-10M is constructed.

Based on Square-10M, we achieve several remarkable results with extensive and rigorous experiments. First, as shown in Figure \ref{img-performance}, our model (TextSquare) achieves comparable or superior performance to advanced closed-source models and substantially outperforms recent state-of-the-art open-source models on various benchmarks. It is notable that the image resolution of TextSquare is $700$ and the parameters are $8.6$B. Second, our experiments validate the beneficial impact of reasoning data on VQA tasks, demonstrating its ability to enhance model performance while mitigating hallucinations. With reasoning data for instruction tuning, TextSquare has a strong reasoning capability to provide elaborate explanations for VQA scenarios. Last but not least, by leveraging the dataset's massive scale, we unveil the relationships between instruction tuning data scale, training convergence loss, and model performance. Whereas a few instruction tuning data can motivate MLLM well, it is not sufficient. Large amounts of high-quality data can still significantly reduce convergence loss and improve performance. The performance of TextSquare grows, and the loss of convergence decreases while continuously scaling up the instruction tuning data, which also demonstrates the effectiveness of our dataset.

In summary, the main contributions of this paper can be categorized into four points:

\begin{itemize}
    \item A high-quality dataset (Square-10M) comprising tens of millions of instances for text-centric VQA instruction tuning is constructed by comprehensively collecting text-rich images from various scenarios and employing the Square (Self-Questioning, Answering, Reasoning, and Evaluation) strategy on closed-source MLLMs.
    \item Leveraging Square-10M, TextSquare achieves a significant outperformance of existing open-source models and even comparable or superior performance to SOTA closed-source models on various benchmarks, e.g., +0.9\% on ChartQA, +2.1\% on WTQ, +4.3\% on SROIE. Notably, TextSquare outperforms GPT4V in overall rankings across ten text-centric benchmarks (ranking 2.2 \textit{v.s.} 2.4).
    \item Reasoning data is demonstrated to be beneficial in improving model performance and mitigating hallucinations in text-centric VQA scenarios, as it can deliver rich question-specific contextual information. 
    \item Through extensive experiments, we reveal the relationships between data scale, convergence loss, and model performance for text-centric VQA instruction tuning, which demonstrates the effectiveness and necessity of Square-10M. 
\end{itemize}

\section{Related Work}

\subsection{Multi-modal Large Language Models}

Recent work has increasingly focused on introducing visual knowledge into LLMs \cite{MLLM-1,MLLM-2,MLLM-3}. General attempts connect a visual encoder and an LLM with intermediate modules like Projector \cite{llava}, Q-Former \cite{blip2}, Perceiver Resampler \cite{flamingo}, etc, and go through pre-training alignment and instruction fine-tuning for vision-language understanding. 

Recently, several researches \cite{Text-MLLM-1,Text-MLLM-2,docpedia,structextv2,vary,omniparser,layoutllm,hrvda} propose to enhance MLLMs' capabilities in understanding textual elements (OCR, text-centric VQA, etc). Among them, mPLUG-DocOwl \cite{Text-MLLM-1} creates novel instruction-following datasets to enhance the tuning process. TextMonkey \cite{MLLM-3} adopts shifted window attention and filters out significant tokens. DocPedia \cite{docpedia} and HRVDA \cite{hrvda} enlarges input resolution to bridge the gap between MLLMs and visual document understanding.

Despite the extraordinary progress of existing open-source MLLMs, they still suffer from the huge gap against SOTA closed-source models like GPT4V \cite{gpt4v} and Gemini Pro \cite{gemini-pro}. In this paper, we propose to mitigate this gap by training with large-scale and high-quality instruction-following data.

\subsection{Text-Centric Visual Question Answering}

Text-Centric Visual Question Answering aims to understand the interactions between the textual and the visual elements in the image. Donut \cite{donut} first proposes an end-to-end training method based on a Transformer without OCR. Pix2Struct \cite{pix2struct} introduces a variable-resolution input representation to adapt to document images. DoCo \cite{doco} enhances the visual representation of the image encoder in LVLMs by aligning the document object of multi-modal inputs. BLIVA \cite{bliva} enlarges the input token space by concatenating learned query embeddings and encoded patch embeddings. Several studies \cite{Text-MLLM-2,tg-doc,llavar} have performed data-centric attempts in this regard. UniDoc \cite{Text-MLLM-2} constructs 600k document-oriented image-text pairs from PowerPoint presentations. LLaVAR \cite{llavar} and TG-Doc \cite{tg-doc} prompt text-only GPT-4 to generate conversations for text-rich images by integrating OCR results into the instructions. These researches are restricted to small-scale annotations or generation based on unimodal inputs.

\subsection{Generating Instruction-Tuning Data via LLMs}

The success of LLMs has inspired recent work to employ them as training data generators \cite{sharegpt4v,allava,self-instruct,synthetic-prompting}. In this regard, we anchor on generating instruction-following data. Self-Instruct \cite{self-instruct} took the initial step towards synthesizing instructions via language models and improving the instruction-following capabilities. Llama-GPT4 \cite{llama-gpt4} uses GPT-4 to generate instruction-following data for LLM fine-tuning. Synthetic Prompting \cite{synthetic-prompting} leverages a few handcrafted examples to prompt LLMs to generate more examples. Bonito \cite{bonito} converts unannotated text into task-specific
training datasets for instruction tuning. Recently, ALLAVA \cite{allava} employs GPT4V to generate reasoning instructions and detailed answers from unlabeled images. All of the above attempts suffer from the low quality of the generated data and are typically performed on a small scale. In contrast, we collect massive text-centric images ({\it i.e.}, tens of millions) and devise comprehensive generating methods and filtering rules to ensure the quantity and quality of the instruction tuning dataset.

\begin{figure}[htbp]
        \centering
\includegraphics[width=\textwidth]{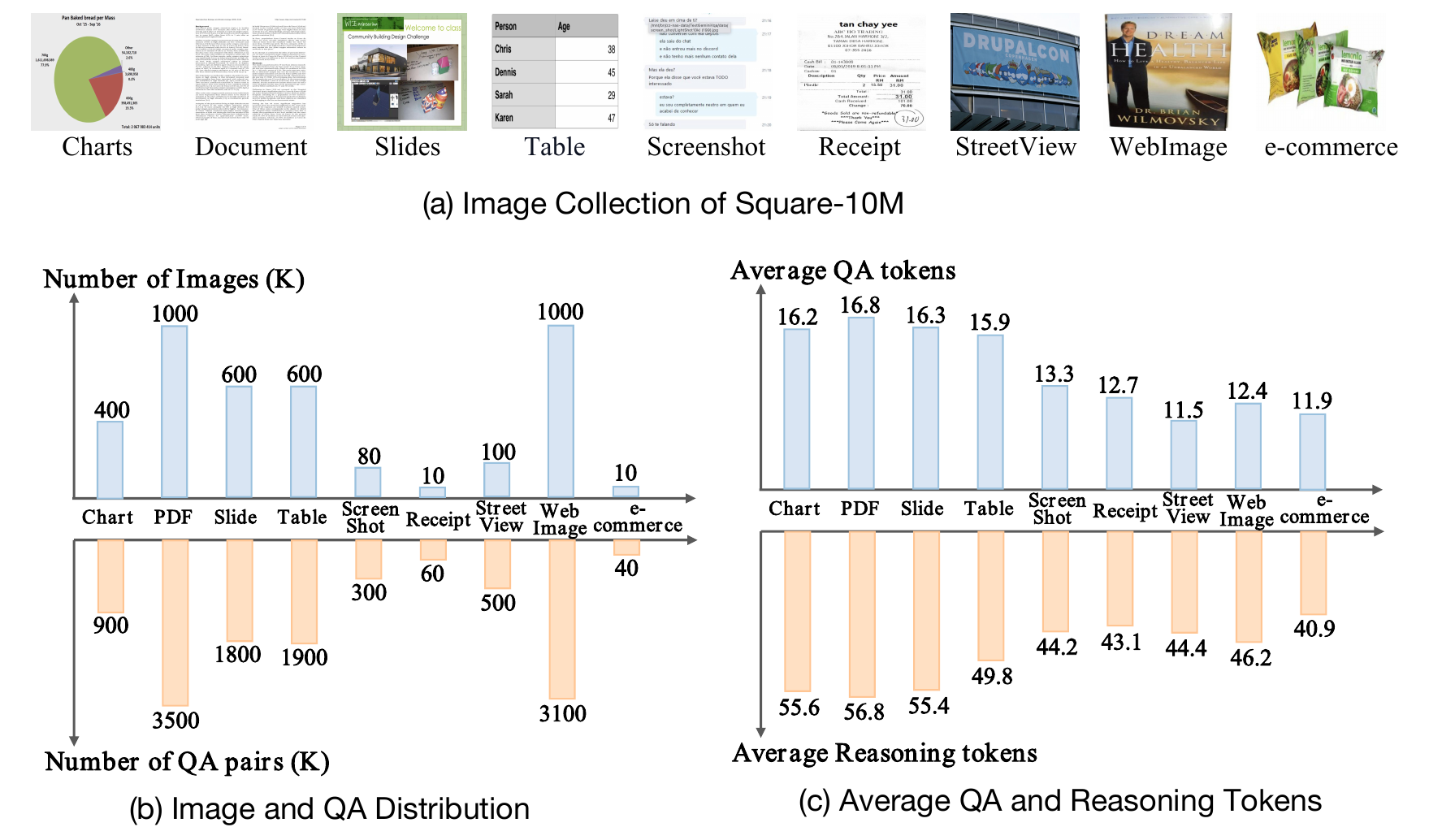}
\caption{Overview of Square-10M: the distribution of images, the average tokens of the QAs, etc.}
\label{data_distribution}

\end{figure}

\section{Square-10M: A Massive and High-quality Text-Centric VQA Instruction Tuning Dataset}

Square-10M is synthesized by our proposed Square pipeline, {\it i.e.}, Self-Questioning, Answering, Reasoning, and Evaluation.

\subsection{Overview of Square}

\begin{figure}[htbp]
        \centering
\includegraphics[width=\textwidth]{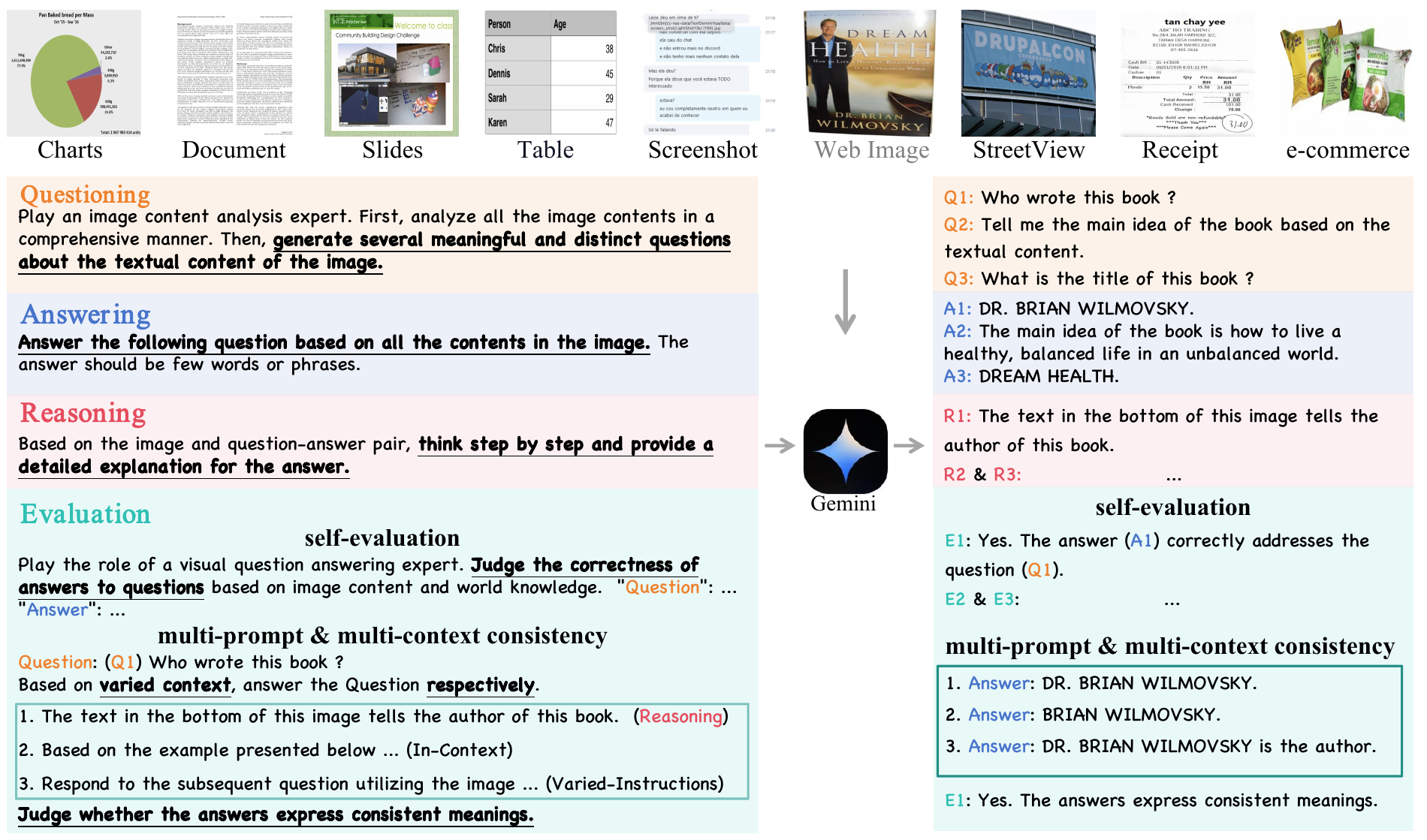}
\caption{Pipeline for the proposed Square strategy. Gemini's versatile multi-modal comprehension capabilities are utilized to synthesize Square-10M, which consists of four stages: self-questioning, answering, reasoning, and evaluation.}
\label{algorithm}
\end{figure}

Figure \ref{algorithm} presents an overview of our proposed Square. Square generally consists of three stages for synthesizing high-quality instruction tuning data for text-centric VQA: (1) Data Collection for collecting large-scale images with textual elements of diverse properties. (2) Data Generation involves self-questioning, answering, and reasoning of the collected data. In this phase, the MLLM is prompted to generate VQA pairs based on the given image, as well as the reasoning behind its answers. (3) Data Filtering for self-evaluation of the generated content, aiming to discard meaningless questions and erroneous answers by employing the evaluation capabilities of MLLMs. 

The above procedures result in our Square-10M dataset, standing out with its massive and high-quality text-centric VQA pairs and reasoning context. To be more specific, a total of 3.8 million images with rich textual elements are collected from diverse sources. After that, 20 million question-answer pairs are obtained from Data Generation. Finally, 9.1 million QA pairs as well as the reasoning context, are distilled with our Square strategy. A more precise analysis of Square-10M is depicted in Figure \ref{data_distribution}.

\subsection{Data Collection}

The data collection strategy is driven by the primary objective of encompassing a broad range of real-world text-rich scenarios. To this end, we collect 3.8 million unlabeled text-rich images (Figure \ref{data_distribution}). These images exhibit diverse properties. For instance, Chart and Table focus on textual elements with intense statistical information; Slide, Screenshot, and WebImage are designed for the interaction between text and prominent visual messages; Document/PDF, Receipt, and e-commerce contain images with fine and dense text; Street-View is derived from natural scenes. The collected images form a mapping of the textual elements in the real world and constitute the foundation of our research on text-centric VQA.

\subsection{Data Generation: Self-Questioning, Answering, and Reasoning}

We build our Square-10M dataset by employing the multi-modal understanding capabilities of Gemini Pro, one of the most advanced LLMs. For each image selected from a specific data source, Gemini Pro is instructed to generate VQA pairs and reasoning context through the subsequent three stages: 

\noindent\textbf{Stage 1: Self-Questioning.}
In this stage, Gemini Pro is prompted to generate profound, meaningful, and non-trivial questions about the given image. We ask Gemini Pro to first comprehensively analyze the image and then raise questions based on its understanding, as shown in Figure \ref{algorithm}. 
Considering that advanced MLLMs typically have a weaker understanding of the textual elements than visual elements, we also prepend the extracted text to the prompt by employing expert OCR models.

\noindent\textbf{Stage 2: Answering.}
Gemini Pro is then instructed to give appropriate answers to the generated questions. We leverage various prompting techniques to enrich the contextual information and improve the reliability of the generated answers, such as Chain-of-Thought and few-shot prompting. Figure \ref{algorithm} shows an example prompt for generating answers to a given question.

\noindent\textbf{Stage 3: Reasoning.}
We require Gemini Pro to elaborate on the detailed reasons behind its answers. Such an effort enforces Gemini Pro to think more about the connections between the questions and the visual elements, thus reducing hallucinations and providing accurate answers. Moreover, the generated reasons could serve as extra contextual information specific to individual questions, favoring possible research on the mechanism behind in-context learning. We present an example prompt for self-reasoning in Figure \ref{algorithm}.

\subsection{Data Filtering: Self-Evaluation and Answering Consistency}
Despite the effectiveness of Self-Questioning, Answering, and Reasoning, the generated image-text pairs could face hallucinatory content, meaningless questions, and erroneous answers. We thus devise filtering rules based on the Evaluation capabilities of LLMs to select high-quality VQA pairs. The whole filtering system is established upon three aspects.

\noindent\textbf{Self-Evaluation of MLLMs.}
We prompt Gemini Pro as well as other advanced MLLMs to judge whether the generated questions are meaningful and whether the answers are good enough to correctly address the questions. 

Figure \ref{algorithm} depicts an example prompt for self-evaluation.

\noindent\textbf{Multi-Prompt Consistency.}
Besides direct evaluation of the generated content, we manually augment the prompt and context space in Data Generation. A correct and meaningful VQA pair should be semantically consistent when provided with different prompts.
Specifically, in the stage of Answering we provide Gemini Pro with different but semantically similar prompts to answer the given question. Then we discard the VQA pairs if the generated answers are not stable in semantics. An example is given in Figure \ref{algorithm}.

\noindent\textbf{Multi-Context Consistency.}
Similar to Multi-Prompt Consistency, we further validate the VQA pairs by prepending the question with varied context information. Given the generated question, three types of answers are produced by Gemini Pro with different contexts: (1) Answering with reasoning. Gemini Pro answers the question with a detailed explanation prepended ({\it i.e.}, content generated in the stage of Reasoning).  (2) In-Context answering. Gemini Pro answers the question with chain-of-thought or few-shot prompts prepended. (3) Naive answering. Gemini Pro answers the question with no extra context.  We then discard the VQA pairs if the generated answers are not semantically consistent.

\section{TextSquare: A Text-Centric Multimodal Large Language Model}

\subsection{Model Architecture}

The model architecture of TextSquare follows the paradigm established by InternLM-Xcomposer2 \cite{internlm-xcomposer2}, including three integral components: (1) A Vision Encoder modified from OpenAI CLIP ViT-L-14-336 \cite{clip}, where the resolution is increased to 700 for improved performance. (2) A LLM based on InternLM-2 \cite{internlm2}, utilizing InternLM2-7B-ChatSFT as the practical variant. (3) A Projector, which semantically aligns the vision token and the text token.

\subsection{Supervised Fine-Tuning with Square-10M}

TextSquare is achieved by performing Supervised Fine-Tuning (SFT) with Square-10 M. The SFT process comprises three stages: In the first stage, we unfreeze all three components ({\it i.e.}, the Vision Encoder, the LLM, and the Projector) and train the model in a resolution of 490. In the second stage, the input resolution is increased to 700, and only the Vision Encoder is trained to adapt to the resolution change. In the third stage, we further perform full-parameter fine-tuning in a resolution of 700. TextSquare demonstrates that with our Square-10M dataset, a model with 8B parameters and normal-size image resolution can achieve extraordinary performance on text-centric VQA, surpassing most available MLLMs and even the closed-source SOTA models.

\section{Experiment}
\subsection{Implementation Details}
The training data contains Square-10M and in-domain datasets (consistent with Monkey's SFT data). The training process is divided into three phases, using the same data and the AdamW \cite{adamw} optimizer with 64 A100-80G GPUs. In the first phase, we fine-tune InternLM-Xcomposer2 with full parameters, and the learning rate decreases from 1e-5 to 1e-6, taking about 9520 GPU hours; In the second phase we scale up the image resolution to 700, and train only VIT, with the learning rate decreasing from 1e-4 to 1e-5, taking about 7280 GPU hours;  In the third stage, we perform full-parameter fine-tuning at 700 image resolution, and the learning rate drops from 1e-5 to 1e-6, spending about 12350 GPU hours.

\subsection{Benchmark Evaluation}
We report the results on Scene Text-centric VQA, Document-oriented VQA, Table VQA, Text-centric KIE, OCRBench, and General VQA for a comprehensive comparison of the performance of our model with existing models. The metrics of each benchmark are listed in Table \ref{benchmark} in the Supplementary Material.

\begin{table}[htbp]
\caption{Quantitative comparison of TextSquare with existing MLLMs on various text-centric benchmarks. ``Res.'' denotes image resolution. ``*'' denotes the results obtained through the open-source checkpoint or API of the closed-source model. The best results of each benchmark are \textbf{bolded}. The best results, except for closed-source models (GPT4V and Gemini Pro), are \underline{underlined.}}
\centering
\setlength\tabcolsep{2pt}
\resizebox{\textwidth}{!}{
\begin{tabular}{l|c|c|ccc|cc|cc|cc}
\hline
\multirow{2}{*}{Method} & \multirow{2}{*}{Res.} & \multirow{2}{*}{OCRBench} & \multicolumn{3}{c|}{Document-Oriented}    & \multicolumn{2}{c|}{Scene Text-Centric} & \multicolumn{2}{c|}{Table VQA} & \multicolumn{2}{c}{KIE}       \\
            &          &         & DocVQA        & ChartQA       & InfoVQA       & AI2D      & TextVQA       & WTQ      & TabFact       & SROIE   & POIE    \\ \hline
UReader \cite{ureader}     & 896  & -    & 65.4   & 59.3  & 42.2  & -   & -   & -   & -     & -    & -    \\
Qwen-VL \cite{MLLM-2}     & 448  & 506  & 65.1   & 65.7  & -     & -   & 63.8        & -     & -       & -     & -  \\
TextMonkey \cite{Text-MLLM-3}  & 896  & 558  & 73.0   & 67.1   &   -  & 44.7   & 65.6     & 37.9  & 53.6       & 46.2    & 32.0    \\
Monkey \cite{monkey}    & 896  & 514  & 66.5   & 65.1  & 36.1  & 57.9$^*$ & 67.6  & 25.3$^*$ & 49.8    & 41.9  & 19.9 \\
Cogagent \cite{cogagent}  & 1120 & 578$^*$  & 81.6   & 68.4  & 44.5  & 49.6$^*$ & \underline{76.1} & 30.2$^*$ & 51.7$^*$  & -  & -  \\
DocOwl 1.5 \cite{docowl-1.5}  & 1344 & 597  & 81.6   & 70.5  & 50.4  & 49.3     & 68.8  & 39.8  & 80.4   & 48.3   & 51.8  \\
Llava Next 34B \cite{llava-next}  & 672  & 573$^*$  & 78.2   & 67.3  & 45.1$^*$  & 70.3   & 69.5    & 47.5$^*$  & 68.9$^*$   & 43.2$^*$   & 46.5$^*$ \\
GPT4V \cite{gpt4v}       & -    & 645  & \textbf{88.4} & 78.5    & 75.1    & 78.2  & \textbf{78.0} & 45.5$^*$ & 69.3$^*$   & 48.9$^*$ & 41.2$^*$ \\
Gemini Pro \cite{gemini-pro}  & -    & \textbf{659} & 88.1    & 74.1   & \textbf{75.2} & 73.9  & 74.6   & 32.3$^*$ & 67.9$^*$   & 38.7$^*$ & 34.6$^*$ \\
Xcomposer2 \cite{internlm-xcomposer2}  & 490  & 511  & 59.6   & 72.7    & 32.9   & 78.7  & 66.1   & 28.7  & 62.3 & 34.2  & 49.3    \\ \hline
TextSquare (ours)  & 700  & \underline{622} & \underline{84.3} & \underline{\textbf{79.4}}  & \underline{51.5} & \underline{\textbf{79.0}} & 66.8 & \underline{\textbf{49.7}}  & \underline{\textbf{84.2}} & \underline{\textbf{53.2}}  & \underline{\textbf{71.8}}  \\ \hline
\end{tabular}
}
\label{table-text-bench}
\end{table}

\textbf{Document-Oriented Benchmark.}
While the documents have a clean background, dense text and complex typography pose distinct challenges. To effectively evaluate our model, we select representative benchmarks including DocVQA \cite{docvqa}, ChartQA \cite{chartqa}, and InfographicVQA \cite{infographicvqa}. The results, detailed in Table \ref{table-text-bench}, show that  TextSquare outperforms all the open-source models in these three document-oriented VQA tasks with an average improvement of $3.5$\%, specifically, DocVQA $84.3$\% \textit{vs.}\ $81.6$\% (Cogagent and mPLUG-DocOwl 1.5), ChartQA $79.4$\% \textit{vs.}\ $72.7$\% (Intern-Xcomposer2), InfographicVQA $51.5$\% \textit{vs.}\ $50.4$\% (mPLUG-DocOwl 1.5). On the ChartQA dataset,  TextSquare outperforms GPT4V and Gemini Pro by a slight margin. Note that TextSquare employs an image resolution of 700, which is smaller than most document-oriented MLLMs. Our model relies on comprehensive, high-quality VQA information specific to the text in the document, improving its ability to recognize and understand various document elements such as text, diagrams, infographics, and so on. If the image resolution is further increased, it is believed that the model performance will be further improved, as demonstrated by Monkey et al.

\textbf{Scene Text-centric Benchmark.}
The ability to answer text-based questions in images becomes an important aspect of the answering task as textual information is usually present in real-world scenes. In the evaluation, we utilize two datasets: TextVQA \cite{textvqa} and AI2D \cite{ai2d}. As shown in Table \ref{table-text-bench}, in this scenario, although TextSquare achieves SOTA performance on the AI2D dataset, there is no major improvement over our baseline Intern-Xcomposer2, which may be due to the fact that Intern-Xcomposer2 has been adequately optimized with high-quality in-domain data.

\textbf{Table VQA Benchmark.} Due to the complex structure of tables and the dense text, the understanding of the content of tables remains a challenging issue. In order to evaluate the performance of the comprehension of table content and structure, we choose two widely utilized datasets, Wiki Table Questions (WTQ) \cite{wtq} and Table Fact (TabFact) \cite{tabfact}, as shown in Table \ref{table-text-bench}. On the Table VQA benchmarks, TextSquare achieves optimal performance among the leading models with an average $3.0$\% improvement. This demonstrates that our model has reached a new level of table understanding, where high-quality generated table VQA and reasoning data play a key role.

\textbf{Text-centric KIE Benchmark.} Text-centric key information extraction tasks are frequently encountered in the information processing of various types of products, certificates, and receipts. We select a receipt information extraction dataset (SROIE) \cite{sroie} and a product information extraction dataset (POIE) \cite{poie}, and the KIE task is converted to the VQA task. TextSquare achieves optimal performance in both datasets, with a major average lift of $14.8$\% (shown in Table \ref{table-text-bench}). It is worth noting that there is no training set of POIE added to the training set, and there is not much data in the domain of product scenarios. This illustrates the extensive textual comprehension capabilities of our model. 

\textbf{OCRBench.} OCRBench \cite{ocrbench} is a comprehensive benchmark consisting of 29 OCR-related assessments, with text recognition, formula recognition, text-centric VQA, KIE, etc. TextSquare achieves optimal performance in OCRBench except for the closed-source models and becomes the first MLLM that exceeds $600$ points with about $10$B parameters. It indicates that the model performs well in both text-centric perception and comprehension tasks, especially in text recognition, where little in-domain data is included in the training set.

\begin{table}[htbp]
\caption{Quantitative comparison of our model with existing MLLMs on representative General VQA and hallucination evaluation benchmarks. VizWiz and POPE are relevant to both VQA and hallucination. Following Cogagent, we evaluate the adversarial part of POPE.} 
\centering
{%
\begin{tabular}{l|cccc|c}
\hline
\multirow{2}{*}{Method} & \multicolumn{5}{c}{General VQA and Hallucination Evaluation} \\ 
                 & VizWiz   & VQAv2    & GQA     & POPE$^{adv}$  & Average \\ \hline
Qwen-VL \cite{MLLM-2}         & 35.2      & 79.5    & 59.3       & -   & -   \\
Monkey \cite{monkey}           & 61.2     & 80.3    & 60.7    & 80.3$^*$  & 70.6   \\
Cogagent \cite{cogagent}         & 36.7$^*$    & \underline{\textbf{83.7}}    & 62.3$^*$   & 85.9   & 67.2    \\
DocOwl 1.5 \cite{docowl-1.5}      & 43.5$^*$    & 68.0$^*$    & 48.5$^*$   & 79.7$^*$   & 59.9  \\
Llava Next 34B \cite{llava-next}   & 63.8     & \underline{\textbf{83.7}}    & \underline{\textbf{67.1}}    & 83.4   & 74.5     \\
GPT4V \cite{gpt4v}           & 64.9$^*$    & 77.2    & 48.4$^*$  & 79.6$^*$  &  67.5 \\
Gemini Pro \cite{gemini-pro}      & 42.8$^*$    & 71.2    & 52.2$^*$   & 84.5$^*$  & 62.7  \\
Xcomposer2 \cite{internlm-xcomposer2}      & 58.9$^*$     & 81.8    & 64.5    & 78.5    & 70.9  \\ \hline
TextSquare (ours)  & \underline{\textbf{71.4}} & 78.0  & 64.5  & \underline{\textbf{86.6}} & \underline{\textbf{75.1}}  \\ \hline
\end{tabular}%
}
\label{table-general-bench}
\end{table}

\textbf{General VQA and Hallucination Evaluation Benchmark.} General VQA requires the ability to learn both visual and textual information and a deep understanding of their inter-relationships. For general VQA, we validate on four benchmarks: VizWiz \cite{vizwiz}, VQAv2 \cite{vqav2}, GQA \cite{gqa}, and POPE \cite{pope}. The VizWiz and POPE benchmarks are also relevant for hallucination evaluation. The results are shown in Table \ref{table-general-bench}. On VQAv2 and GQA, TextSquare does not have a significant degradation compared to InternLM-Xcomposer2 and still maintains comparable performance. TextSquare exhibits superior capabilities in VizWiz and POPE, outperforming the closest competing method by an average of $3.6$\%. These results highlight the effectiveness of our approach, which is also able to mitigate model hallucinations in particular with large-scale instruction tuning. We observe that it is partly attributed to the high-quality reasoning data that provides detailed explanations for VQA.

\subsection{Qualitative Analysis}
\begin{figure}[htbp]
\centering
\includegraphics[width=\textwidth]{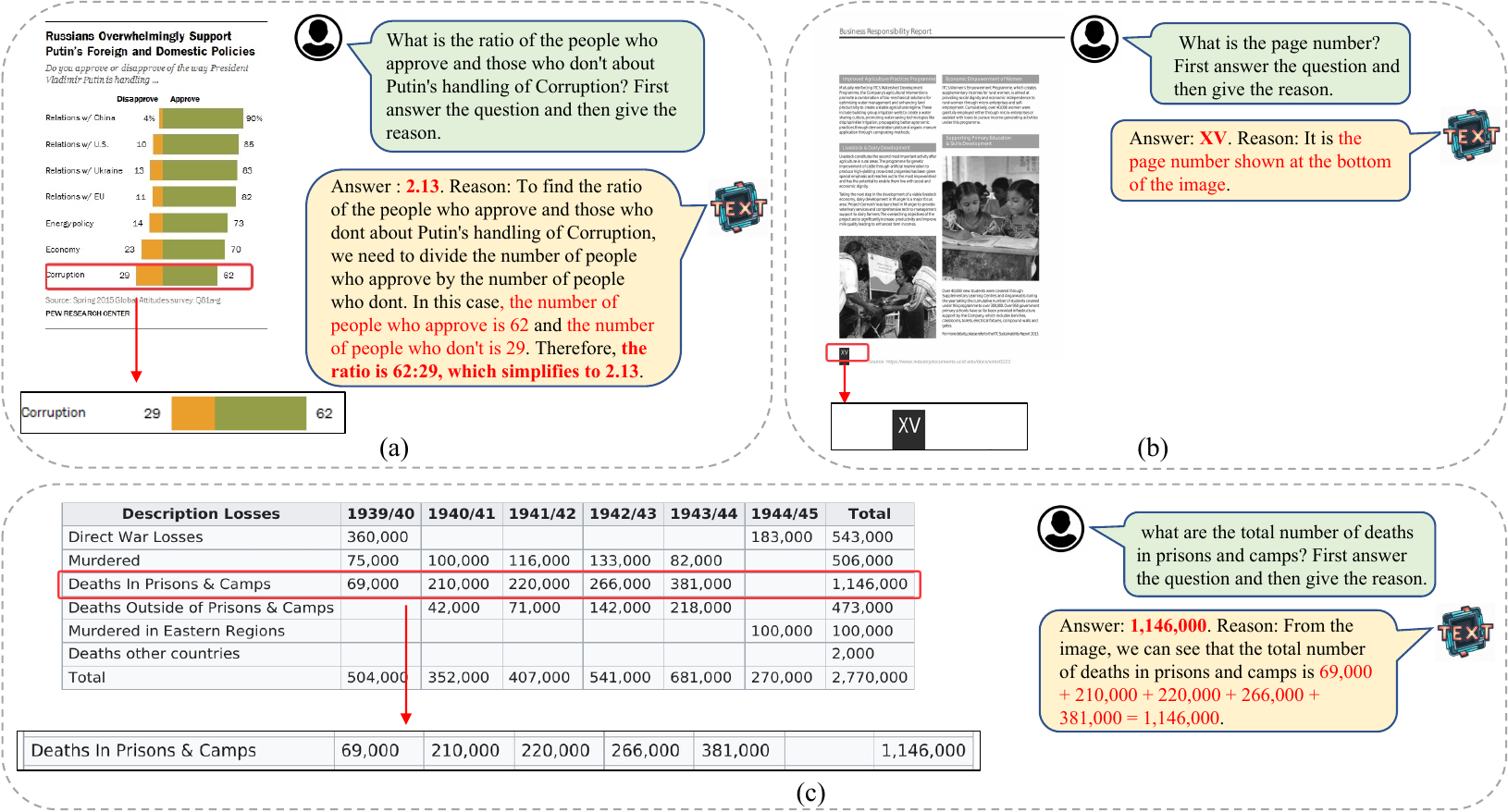}
\caption{Qualitative results of VQA and reasoning for various text-centric scenarios.}
\label{reasoning_case}
\end{figure}

As illustrated in Figure \ref{reasoning_case}, TextSquare has a formidable capability to provide plausible explanations of the answers to questions in a variety of text-centric VQA scenarios. Figure \ref{reasoning_case}(a) shows that TextSquare has simple arithmetic capabilities. Figure \ref{reasoning_case}(b) shows the ability to understand textual content and provide an approximate location in dense text.
Figure \ref{reasoning_case}(c) shows the comprehension of table structure and the ability to extract contextual information relevant to the question.

\begin{table}[htbp]
\centering
\caption{Ablation study on Incorporating Square-10M for Instruction Tuning.}
\begin{tabular}{l|c|ccccc|c}
\hline
Model & OCRBench & DocVQA & ChartQA & InfoVQA & WTQ & SROIE & Average  \\ \hline
Xcomposer2$^*$  & 571 &74.8 & 73.2 & 41.6 & 40.3 & 44.7 & 54.9 \\
TextSquare      & 622 &84.3 &79.4 & 46.2 & 49.7 & 53.2 & 62.6 \\
\hline
\end{tabular}
\end{table}

\begin{table*}
\resizebox{0.95\textwidth}{!}{
\begin{floatrow}
\capbtabbox{
\setlength\tabcolsep{2pt}
\begin{tabular}{c|ccc}
\hline
Evaluation & DocVQA & ChartQA & WTQ  \\ \hline
w/    & 84.3   & 79.4    & 49.7 \\
w/o   & 81.7   & 77.2    & 46.9 \\
\hline
\end{tabular}
}{
 \caption{Ablation study on the evaluation step in the Square strategy.}
 \label{Tab1}
}
\capbtabbox{
\setlength\tabcolsep{2pt}
\begin{tabular}{c|cccc}
\hline
Reasoning Data & DocVQA & ChartQA & POPE$^{adv}$ & WizViz \\ \hline
w/   & 84.3   & 79.4    & 86.5 & 71.4   \\
w/o  & 82.9   & 78.1    & 83.8 & 68.2  \\
\hline
\end{tabular}
}{
 \caption{Ablation study on the VQA Reasoning data of Square-10M.}
 
 \label{Tab2}
 \small
}
\end{floatrow}
}
\end{table*}

\subsection{Ablation Study}

\textbf{The Effect of Incorporating Square-10M for Instruction Tuning.} 

In order to verify the effectiveness of Square-10M, we fine-tune the baseline model InternLM-Xcomposer2 on the public text-centric VQA instruction tuning dataset (consistent with Monkey's training data). As shown in the Table, TextSquare substantially outperforms Xcomposer2$^*$ (fine-tuned) on various text-centric VQA benchmarks by $7.7$\%, which corroborates that Square-10M can fully exploit MLLM's ability in text-centric VQA scenarios and that a large amount of high-quality instruction tuning data has a major improvement in performance.

\textbf{The Effect of the Evaluation Step of the Square Strategy.} As shown in Table \ref{Tab1}, there is a distinct improvement in model performance after incorporating the evaluation of the generated VQA data, which verifies that the evaluation step of the Square strategy improves the quality of VQA instruction tuning data.

\textbf{The Effect of VQA Reasoning Data on Model Performance and Hallucination Evaluation.} From Table \ref{Tab2}, we can find that VQA Reasoning data is helpful in both improving VQA performance and mitigating hallucinations. Specifically, in terms of enhancing VQA performance, there is a 1.4\% and 1.3\% gain on DocVQA and ChartQA. In terms of mitigating hallucinations, there is a $2.7$\% and $3.2$\% gain on POPE and WizViz.

\begin{figure}[htbp]
\centering
\includegraphics[width=0.95\textwidth]{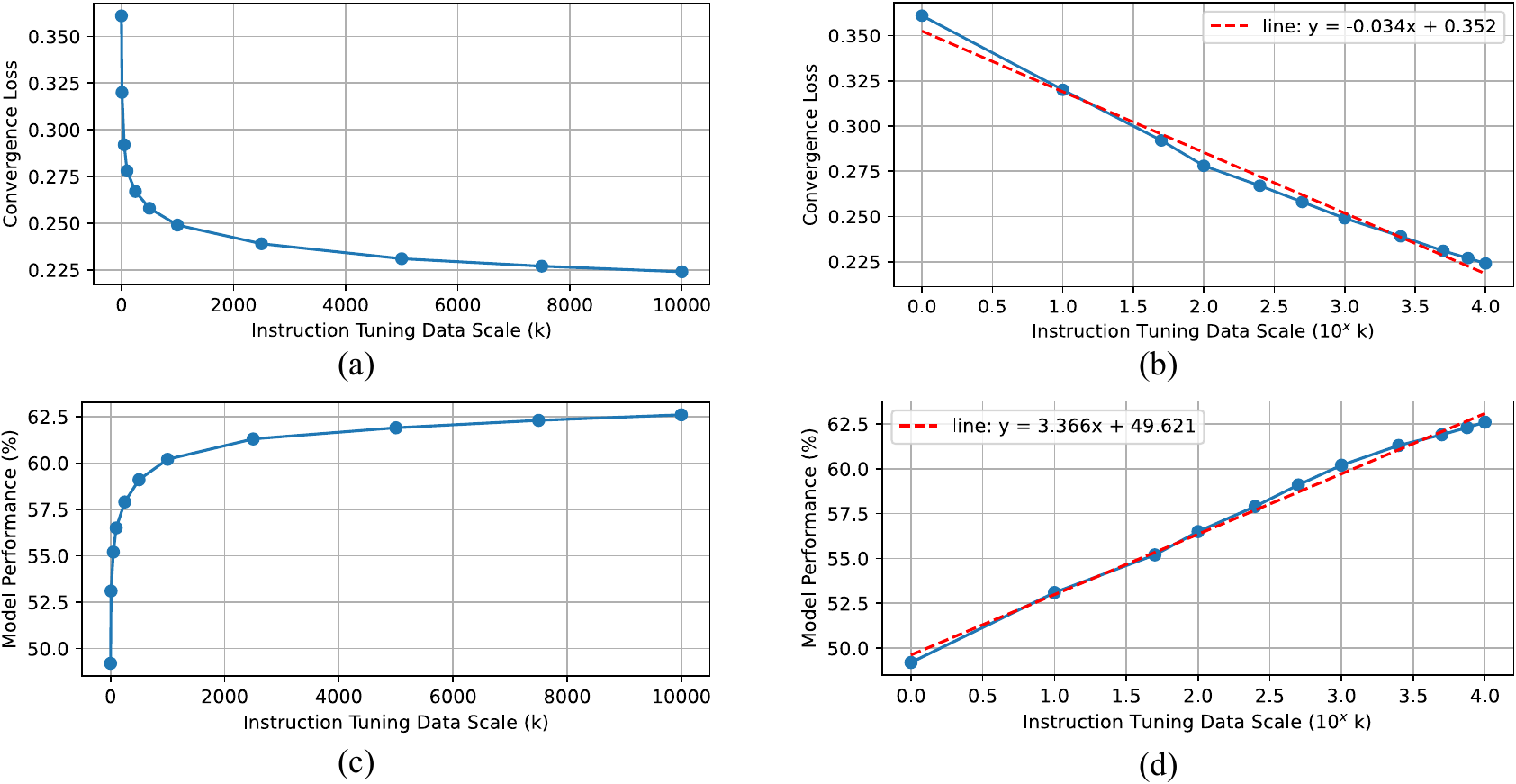}
\caption{The relationship between instruction tuning dataset scale, convergence loss, and model performance in text-centric VQA scenarios. Figure (a) and Figure (b) show the relationship between data scale and convergence loss, distinguished by a scaling of the horizontal coordinate of Figure (b) with log$_{10}$. Figures (c) and (d) show the relationship between data scale and model performance, distinguished by a scaling of the horizontal coordinate of figure (e) with log$_{10}$.}
\label{data-scale}
\end{figure}

\subsection{Relationships between Instruction Tuning Data Scale, Convergence Loss, and Model Performance}

To explore the relationship between instruction tuning data scale, convergence loss, and model performance based on the merged large-scale Square-10M and the in-domain instruction tuning dataset, we conduct 10 sets of experiments for different data scales. The average performance of the models is evaluated on DocVQA, ChartQA, InfoVQA, WTQ, and SROIE. As shown in Figure \ref{data-scale}(a)(b), the convergence loss of the model continues to decrease as the data scale grows, whereas the rate of decrease becomes progressively slower. The relationship between the convergence loss and the instruction tuning data scale approximately conforms to a logarithmic function. Similarly, from Figure \ref{data-scale}(c)(d), it can be seen that as the instruction tuning data grows, the model performs better and better, but the rate of growth continues to slow down. Their relationship is also approximately in accordance with a logarithmic function. Holistically, there is a corresponding scaling law in the instruction tuning phase in text-centric VQA scenarios, where model performance is proportional to the logarithm of the scale of data. It can guide the construction of potentially larger datasets and predict model performance.

\section{Limitation}
Although our approach achieves remarkable results in various scenarios, there are some limitations. Firstly, large-scale data requires plenty of GPUs for long-term training, which greatly increases the training consumption. Second, while the Square strategy improves the quality of synthetic data, it still cannot reach the human level.

\section{Conclusion}
In this paper, we present the Square strategy for constructing a high-quality text-centric instruction tuning dataset(Square-10M). 
Leveraging this dataset, TextSquare significantly surpasses recent open-source models and even achieves performance comparable to GPT4V across various benchmarks.
Furthermore, we derive the relationship between instruction tuning dataset scale, convergence loss, and model performance in order to pave the way for constructing even larger datasets. Our approach provides a data-centric perspective that revisits the role of instruction-tuning data in text-centric VQA, confirming that both the quantity and quality of data are crucial to model performance. We believe that there is a promising direction on how to further improve the data quantity and quality for closing the gap between open-source models and the leading ones.

\newpage

{
\small

\bibliographystyle{plainnat}
\bibliography{main}

}

\newpage

\section{Supplementary Material}
\subsection{Summary of the Evaluation Benchmarks}
We summarize the evaluation benchmarks used in this paper in Table \ref{benchmark}.

\begin{table}[htbp]
\centering
\caption{Summary of the evaluation benchmarks.}
\setlength\tabcolsep{2pt}
\resizebox{\textwidth}{!}{
\begin{tabular}{l|l|l|l}
\hline
Benchmark & Description                                  & Split & Metric    \\ \hline
DocVQA    & VQA on document images                       & test  & ANLS      \\ \hline
ChartQA & VQA on charts with visual and logical reasoning                  & test              & Relaxed Accuracy \\ \hline
InfoVQA   & VQA on infographic images                    & test  & ANLS      \\ \hline
AI2D      & Multiple choice VQA on science diagrams      & test  & Accuracy  \\ \hline
TextVQA & VQA involving reading and reasoning about text                   & val               & VQA Score        \\ \hline
WTQ     & VQA on semi-structured HTML tables sourced from Wikipedia        &test                   & Accuracy         \\ \hline
TabFact   & 'Yes' or 'No' choice VQA about tables        & test  & Accuracy  \\ \hline
SROIE     & Key information extraction  from receipts    & test  & Accuracy  \\ \hline
POIE      & Key information extraction on product images & test  & Accuracy  \\ \hline
VizWiz    & Answering visual questions from blind people & val   & VQA Score \\ \hline
VQAV2     & Open-ended VQA about natural images          & val   & VQA Score \\ \hline
GQA     & Real-world visual reasoning and compositional question answering & test-dev          & Accuracy         \\ \hline
POPE    & Yes-or-No VQA to assess the object hallucination problem         & test(adversarial) & F1 Score      \\ \hline
\end{tabular}
}
\label{benchmark}
\end{table}

\end{document}